\documentclass[journal]{IEEEtran}

\newcommand{\revised}[1]{{\color{black}#1}}

\usepackage{etoolbox}

\usepackage{amsmath}
\usepackage{algorithm}
\usepackage{algpseudocode}
\usepackage{booktabs}
\usepackage{multirow}
\usepackage[utf8]{inputenc} 
\usepackage[T1]{fontenc}    
\usepackage{hyperref}       
\usepackage{url}            
\usepackage{booktabs}       
\usepackage{amsfonts}       
\usepackage{nicefrac}       
\usepackage{microtype}      
\usepackage{xcolor}         
\usepackage{graphicx}
\usepackage{float}
\usepackage[numbers,sort&compress]{natbib}

\newfloat{figtab}{htb}{fgtb}
\makeatletter
  \newcommand\figcaption{\def\@captype{figure}\caption}
  \newcommand\tabcaption{\def\@captype{table}\caption}
\makeatother

\def\BibTeX{{\rm B\kern-.05em{\sc i\kern-.025em b}\kern-.08em
    T\kern-.1667em\lower.7ex\hbox{E}\kern-.125emX}}

\title{A Fuzzy Logic-based Approach to Predict Human Interaction by Functional Near-Infrared Spectroscopy}

\author{
\IEEEauthorblockN{Xiaowei Jiang\textsuperscript{\dag}, Liang Ou\textsuperscript{\dag}, Yanan Chen, Na Ao, Yu-Cheng Chang, Thomas Do, Chin-Teng Lin\textsuperscript{*}}

\thanks{\textsuperscript{\dag}Xiaowei Jiang and Liang Ou contributed equally to this work.}
\thanks{Xiaowei Jiang, Liang Ou,  Yu-Cheng Chang, Thomas Do, and Chin-Teng Lin are with the GrapheneX-UTS Human-centric AI Centre, Australian AI Institute, School of Computer Science, Faculty of Engineering and Information Technology, University of Technology Sydney.}
\thanks{Yanan Chen and Na Ao are with the Institute of Psychology and Behavior, Henan University.}

\thanks{\textsuperscript{*}Corresponding author: Chin-Teng Lin. Email: chin-teng.lin@uts.edu.au}
}

\begin{document}
\maketitle


\begin{abstract}
This paper introduces the Fuzzy logic-based Attention (Fuzzy Attention Layer) mechanism, a novel computational approach designed to enhance the interpretability and efficacy of neural models in psychological research. 
The Fuzzy Attention Layer integrated into the Transformer Encoder model to analyze complex psychological phenomena from neural signals captured by functional Near-Infrared Spectroscopy (fNIRS).
By leveraging fuzzy logic, the Fuzzy Attention Layer learns and identifies interpretable patterns of neural activity. This addresses a significant challenge in using Transformers: the lack of transparency in determining which specific brain activities most contribute to particular predictions. Our experimental results, obtained from fNIRS data engaged in social interactions involving handholding, reveal that the Fuzzy Attention Layer not only learns interpretable patterns of neural activity but also enhances model performance. Additionally, these patterns provide deeper insights into the neural correlates of interpersonal touch and emotional exchange. The application of our model shows promising potential in understanding the complex aspects of human social behavior, verify psychological theory with machine learning algorithms, thereby contributing significantly to the fields of social neuroscience and AI.
\end{abstract}

\begin{IEEEkeywords}
Fuzzy Logic, Transformers, fNIRS, Social Neuroscience
\end{IEEEkeywords}
\section{Introduction}

\begin{figure*}[htbp]
    \centering
    \includegraphics[width=\linewidth]{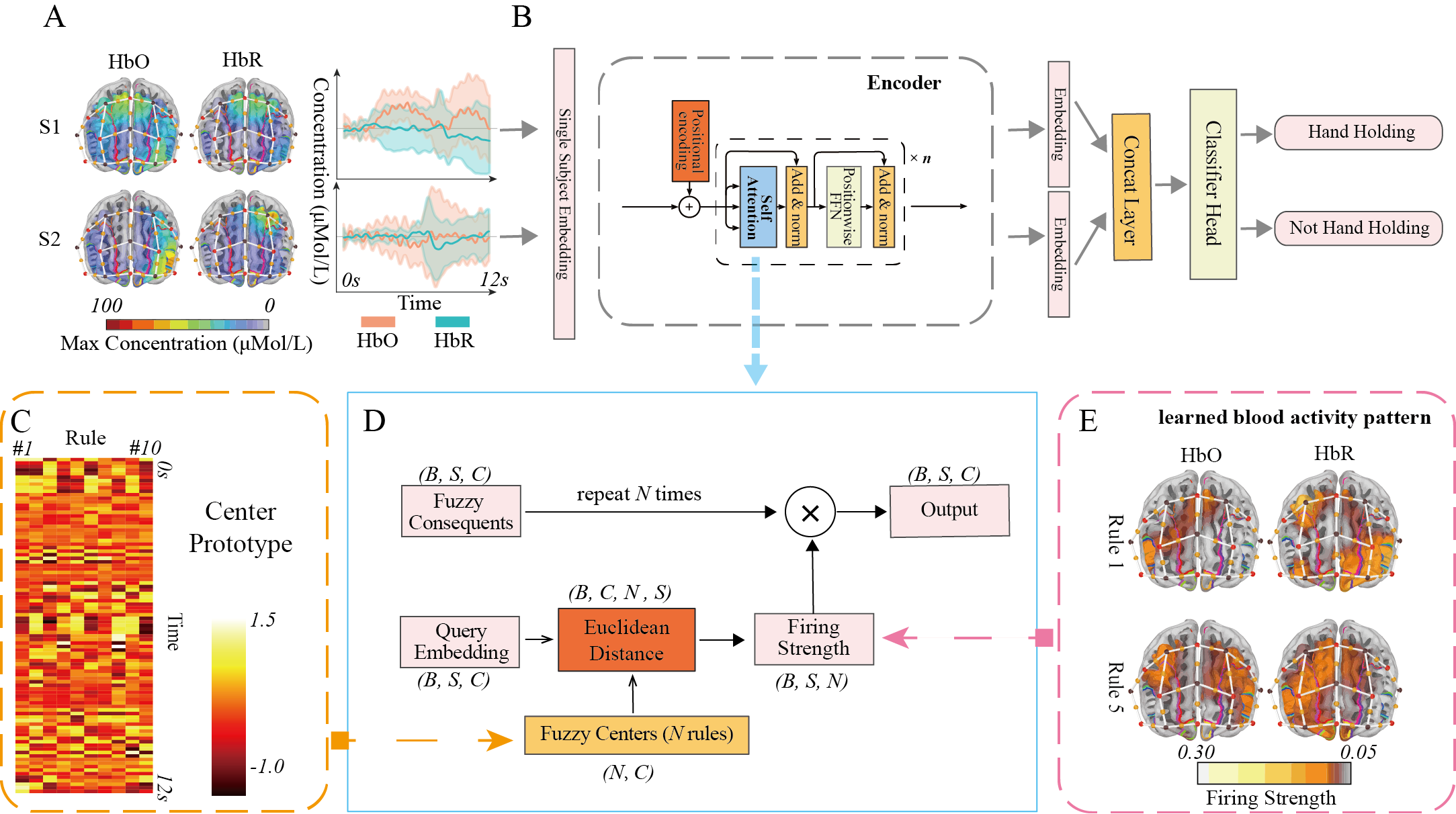}
    \caption{Overall illustration of decoding handholding by hyperscanning fNIRS signals. \textbf{A}: A demonstration sample showing paired fNIRS signals. The brain image displays the prefrontal cortex's maximum concentration of HbO and HbR. The second plot illustrates how the average signal across channels for HbO and HbR changes over time. \textbf{B}: The main structure of our proposed model. Add\&Norm: A residual connection followed by layer normalization. \textbf{C}: The center, which is identified by the Fuzzy Attention Layer. \textbf{D}: The structure of the Fuzzy Attention Layer. \textbf{E}: The firing strength of the demonstration sample.}
    \label{fig:Sample_Explain}
\end{figure*}

\IEEEPARstart{H}{uman} Interaction Behaviour emerges at the confluence of psychology, engineering and artificial intelligence, marking a developing yet pivotal field aimed at decoding the intricate dynamics of human social behaviors. This domain traditionally harnesses psychological and neuroscientific theories to explore the nuances of individual interactions and emotional bonds \cite{farroniSelfregulatoryAffectiveTouch2022, packheiserSystematicReviewMultivariate2024, gazzolaPrimarySomatosensoryCortex2012, liNeuralSynchronizationPredicts2022, hyonSocialNetworkProximity2020, parkinsonSimilarNeuralResponses2018, suvilehtoWhyWhoHow2023} and be applied to robotics \cite{guoHowRobotTouch2024}. In this context, a focal point of non-verbal communication is interpersonal touch \cite{farroniSelfregulatoryAffectiveTouch2022, packheiserSystematicReviewMultivariate2024}, specially handholding, which originates as the foremost sensory channel during prenatal development \cite{trevarthen1993perceived} and persists as a crucial medium for expressing intentions and emotions in adulthood \cite{goldsteinBraintobrainCouplingHandholding2018, coanMutualitySocialRegulation2013}. The profound impact of touch is evidenced in its essential role across diverse social functions—ranging from flirtation and dominance assertion to comfort and caregiver-child bonding—thereby enriching social exchanges with emotional depth and fostering feelings of support, calmness, trust, and security among close individuals \cite{VALORI2024108121}. 

Despite its rich theoretical reinforcements, the field has primarily employed methods with limited capacity to validate imperfect theoretical claims, thus presenting a significant challenge in demonstrating foundational assumptions \cite{varoquaux2021ai, farahaniIntroductionArtificialPsychology2023, challenge2019}. Explainable AI (xAI) has recently been developed to elucidate the results of supervised machine learning models \cite{GunningXAI}. The advent of xAI introduces innovative methods for analyzing and interpreting high-dimensional data, such as neural responses, offering a promising avenue for integrating psychological theories and neuroscientific insights within social cognitive and affective neuroscience—a venture that remains largely untapped.

Several popular xAI methods include perturbation-based approaches \cite{IVANOVS2021228}, which perturb features and observe prediction changes, and Local Interpretable Model-Agnostic Explanations (LIME) \cite{10.1145/2939672.2939778}, which analyzes changes in individual predictions based on input variations. LIME generates interpretable indications of each feature's contribution by controlling the feature value in a single input. Deep Learning Important FeaTures (DeepLIFT) \cite{shrikumar2017just} explains output predictions by back-propagating neuron contributions in a neural network, using a partial derivative-like function with the chain rule to calculate feature importance scores. SHAP (SHapley Additive exPlanations) \cite{lundberg2017unified} uses Shapley values from game theory to explain model outputs, detailing each feature's contribution. DeepSHAP \cite{10190801} combines SHAP values and DeepLIFT to assess feature significance through linear composition rules and backpropagation. However, these methods have drawbacks, such as sensitivity to minor changes causing instability \cite{alvarez2018robustness} and high computational complexity.

In neuroscience and psychology, explaining a theory involves elucidating the underlying mechanisms and processes that account for observed behaviors and phenomena. Traditional methods, like t-tests or ANOVA, can describe differences between groups, while the General Linear Model (GLM) is another popular approach for analyzing neural signals. However, these methods cannot accurately describe the data at an individual level, as they only yield group-level results. To overcome these limitations and achieve more precise and individualized insights, advanced techniques such as machine learning and xAI are increasingly being utilized \cite{10.1145/3351095.3375624, BADRULHISHAM2024470, holzinger2017need}. For instance, EEGNet \cite{Lawhern_2018} designs a CNN-based model and explains frequency and space features using kernels in different layers. Another example is based on fNIRS \cite{shibuExplainableArtificialIntelligence2023}, which combines CNN and LSTM as feature extractors and then explains features using SHAP \cite{lundberg2017unified}. However, these methods are limited because they cannot simultaneously describe both sample-specific and global features and are designed for specific signals. The challenge remains in developing a method that comprehensively describes the underlying mechanisms in general tasks at both the global and individual levels.

This paper introduces an interdisciplinary, bottom-up approach with an explainable fuzzy logic-based model to support psychological and neuroscientific theories. We propose the Fuzzy logic-Based Attention (Fuzzy Attention Layer), illustrated in \revised{Fig.} \ref{fig:Sample_Explain}D.

Designed for data-sparse environments like functional Near-Infrared Spectroscopy (fNIRS) \cite{Ghosh2022,8467365}, it automatically learns the neural activity patterns in a human-interpretable manner, providing evidence for human interaction behavior and enhancing performance compared to traditional dot-product attention in neural datasets.

The Fuzzy Attention Layer combines Fuzzy set theory \cite{shihabudheen2018recent}, Fuzzy neural networks \cite{10183374,106218}, and the Transformer sequence modeling \cite{NIPS2017_3f5ee243}, modeling input sequences as fuzzy sets that output firing strengths as an $S$ by $N$ matrix, indicating rule recognition per token. This approach not only parallels the mechanics of vanilla dot-product self-attention \cite{cheng2016long, NIPS2017_3f5ee243, paulus2018a} but also incorporates fuzzy logic, making it apt for fNIRS data analysis\cite{KarmakarFuzzyfNIRS, BEZDEK1984191}. Our experiments show that this layer offers superior approximation capabilities, helping elucidate data patterns and underlying reasoning.

This article is structured as follows. In Section II, we review recent work on brain decoding in human interactions and fuzzy inference systems. Section III describes the design and implementation of the proposed fuzzy attention layer and the fuzzy Transformer model. Experimental results and ablation studies are presented in Section IV to demonstrate the performance of the proposed approach and provide insights into the model’s behavior. Finally, conclusions and potential future research directions are discussed in Section V.

The contributions of this study include:
\begin{enumerate}
    \item Proposing Fuzzy Attention Layer: The first to design an attention kernel based on fuzzy theory to improve the interpretability of a Transformer without sacrificing its performance. 
    \item Empowering of the Transformer Encoder: The integration of the Fuzzy Attention Layer enables the Transformer Encoder to effectively capture fNIRS input patterns and their inter-channel interactions. 
    \item Demonstrating Interpretability Analysis: By conducting a thorough interpretability analysis, the learned fuzzy rules are leveraged to assess individual sample contributions and the overall decision-making process.
    \item Revealing underlying Neural Patterns: By examining the fuzzy rules learned by the Fuzzy Attention Layer, we uncover the underlying neural patterns of specific human-to-human interactions , and it connects machine learning algorithms with psychological theories.
\end{enumerate}

\begin{figure*}[htbp]
    \centering
    \includegraphics[width=\linewidth]{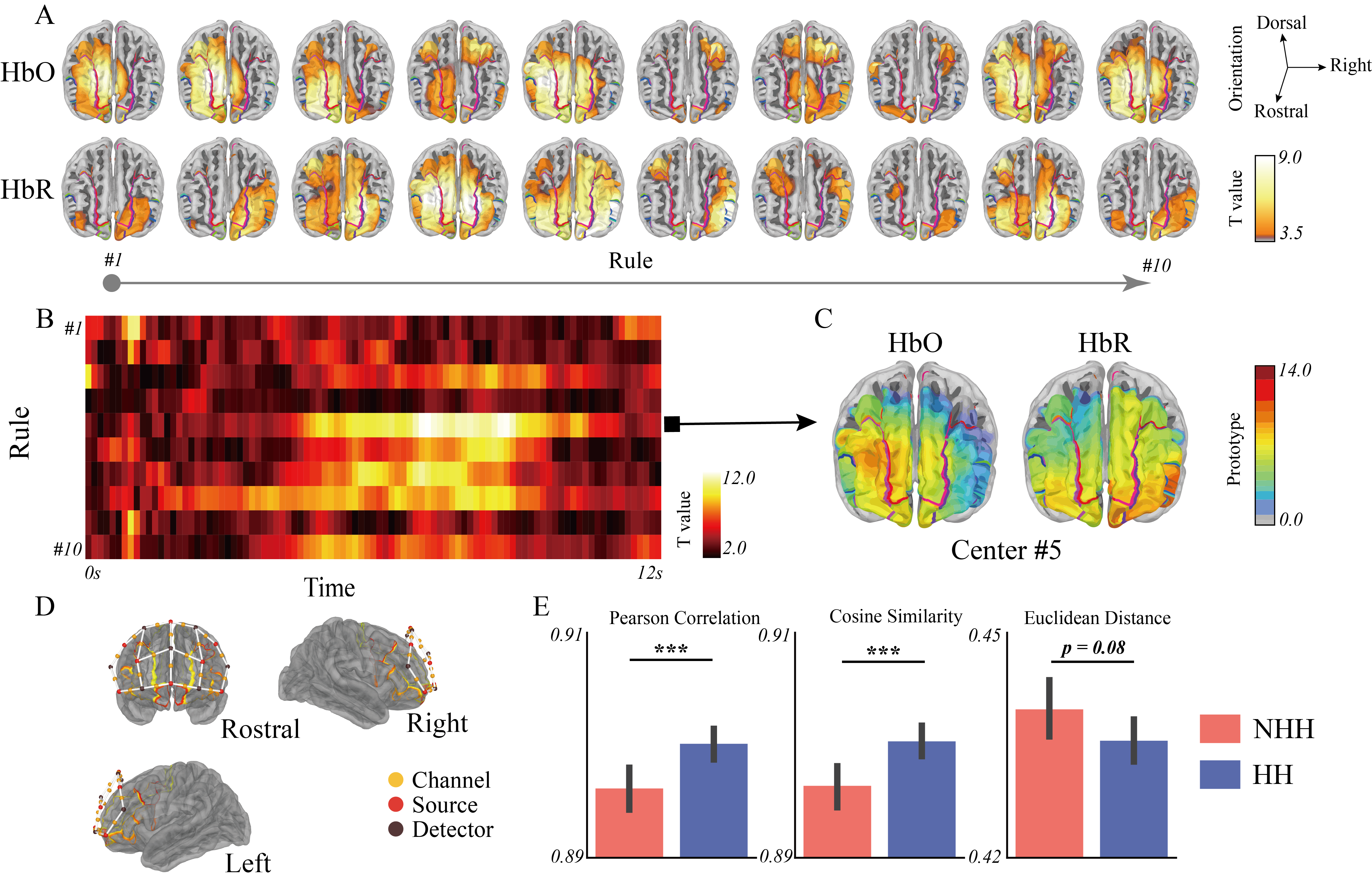}
    \caption{
    \revised{\textbf{A}: The T-values, which illustrate the differences between handholding (HH) and non-handholding (NHH) conditions across 10 different rules, are compared within the channel-first structure. }
    \revised{\textbf{B}: The T-values, which illustrate the differences between HH and NHH conditions across 10 rules, are compared within the time-first structure. }
    \revised{\textbf{C}: The centre prototype for rule \#5.}
    \textbf{D}: The placement of fNIRS sensors (sources and detectors) and channels. \textbf{E}: Statistical analysis results showing the (de-)similarity between the latent variables of two individuals. ***:\textit{p}<0.001}
    \label{fig:Group_rule.png}
\end{figure*}

\section{Related Work}

\subsection{Brain Decoding in Human Interactions}
\label{sec:Related_Work_Human_Interactions_Brain_Decoding}

Over the past decade, significant advances have been made in brain decoding, particularly in the realm of human interactions. Research in this area generally falls into two categories: verbal and non-verbal communication. Earlier studies focused on the evolution of spoken language, exploring the connection between semantic features and neural activities \cite{huthNaturalSpeechReveals2016}. Pioneering work by Bartlett revealed the role of acoustic features in speech perception \cite{bartlettOrganizationPhysiologyAuditory2013}. This was followed by Chang et al., who demonstrated how the superior temporal gyrus (STG) encodes phonetic features \cite{mesgaraniPhoneticFeatureEncoding2014}. Detailed examinations of semantic processing patterns in the human inferior frontal gyrus were conducted by Wu et al. \cite{zhuDistinctSpatiotemporalPatterns2022}. Studies have also explored how humans acquire vocal and spoken language, linking it to emotional memory \cite{jarvisEvolutionVocalLearning2019, buchananVerbalNonverbalEmotional2001}. Willett et al. developed intra-cortical BCI systems that serve as speech neuroprostheses \cite{willettHighperformanceSpeechNeuroprosthesis2023} and decode handwriting into text \cite{willettHighperformanceBraintotextCommunication2021}, utilizing RNNs. Furthermore, Wu et al. introduced methods for decoding tonal language speech \cite{liuDecodingSynthesizingTonal2023}. Notably, fNIRS has also been evaluated for its ability to decode brain activity, reflecting semantic information triggered by natural words and concepts \cite{caoBrainDecodingUsing2021}. In non-verbal communication, research remains sparse, focusing primarily on theoretical aspects. Some studies have shown that Inter-Brain Synchrony (IBS) can predict social relationships, such as friendships \cite{parkinsonSimilarNeuralResponses2018} and romantic partnerships \cite{zhangAffectiveCognitiveInterpersonal2023}, as well as emotions like marital satisfaction \cite{liNeuralSynchronizationPredicts2022}. To investigate these phenomena, simultaneous brain signal recording, known as the hyperscanning approach, is essential \cite{babiloniSocialNeuroscienceHyperscanning2014, goldsteinBraintobrainCouplingHandholding2018, dumasSocialBehaviourBrain2011, czeszumskiHyperscanningValidMethod2020}. Recently, the field has shown increased interest in how touch influences neural responses. Packheiser et al. summarized the benefits of touch interventions \cite{packheiserSystematicReviewMultivariate2024}, noting that social touch evokes a complex network of social-affective reactions in the brain \cite{leemassonRapidProcessingObserved2023}. Alouit et al. described the spatiotemporal dynamics of tactile information remapping in the brain \cite{alouitCorticalActivationsAssociated2024}. Additionally, fNIRS, as a standard hyperscanning device,  has been applied in novel contexts, from studying infant behavior and maternal sensitivity \cite{roche-labarbeCoupledOxygenationOscillation2007, hydeNearinfraredSpectroscopyShows2010, nguyenProximityTouchAre2021, mateusMaternalSensitivityInfant2021} to investigating emotion and social regulation in human cognition \cite{zhangAffectiveCognitiveInterpersonal2023, coanMutualitySocialRegulation2013}. Remarkably, it has also been employed in robotics to assess users' emotional responses \cite{guoHowRobotTouch2024}.

\subsection{Fuzzy Inference Systems}

This paper design the FIS based on the Takagi–Sugeno–Kang (TSK) inference system, which composed of several IF-THEN rules, as follows:

\begin{equation}
\mathrm{If}\ x\ \mathrm{is}\ \mu_r,\ \mathrm{Then\ the\ output\ is}\ u_{r},
\label{eq.rule}
\end{equation}

\noindent where $x$ denotes the input variable, $\mu_r$ is the firing strength of fuzzy set $r$, and $u_{r}$ is the output of rule $r$. The $\mu_r$ in TSK is calculated using the product of Gaussian membership functions:
\begin{equation}
\mu_r\left(x\right)=\prod_{d}\exp\left\{-\frac{1}{2 }\left(\frac{x_d-m_{d,r}}{\sigma_{d,r}   }\right)^2\right\},
\label{eq.fs}
\end{equation}
where $m_{d,r}$ and $\sigma_{d,r}$ are the centre and width of fuzzy set $r$. The output $u_{r}$ can be a single value or a linear projection from $x_d$. The final output of TSK FIS is:
\begin{equation}
a=\sum_{j}^{R}{\frac{\mu_j(x)u_j}{\sum_{i}^{R}{\mu_i(x)}}}, 
\label{eq.rule_out}
\end{equation}
where $R$ is the number of rules, and $a$ is a summation of all rules' weighted outputs after normalization.

FISs have been further developed into a neural network architecture known as the Fuzzy Neural Network (FNN) \cite{lin1996neural}, which can be trained using gradient descent optimization. The advantage of this architecture is that the fuzzy logic follows human intuitive deduction, providing good explainability. Additionally, the nonlinearity provided by the Gaussian membership functions makes the approximation of real-world data more robust \cite{zhang2023robust}. A recent study, KAN \cite{liu2024kan}, suggests that by learning the activation function for each dimension, the network provides "internal degrees of freedom". This freedom is modeled in TSK through $m_{d,r}$ and $\sigma_{d,r}$, shaping different bell-shaped activations for each $x_d$. This is entirely different from linear-based models (Transformers and CNNs), where each $x_d$ is only multiplied by a weight. It is verified in this paper that the Fuzzy Attention Layer achieves more extraordinary approximation ability by providing internal degrees of freedom.

From the perspective of brain signal processing, adapting FISs has been proven effective in many studies \cite{chang2021exploring,reddy2021joint,hu2021fuzzy,zarandi2011systematic}. These studies combine FISs with state-of-the-art machine learning models and achieve impressive improvements. However, such combinations are often just parallel or serial connections, such as an FIS after a deep CNN or an FIS beside a Transformer attention block. The disadvantage is that deep learning does not integrate the FIS advantages but merely mixes them.

In this study, we take a further step by combining attention calculation with FIS, thus proposing the Fuzzy Attention Layer. \revised{Details of this implementation is discussed in the next section.}


\section{Methodology}

\subsection{Task Definition}
To explore the participants' interaction state, the experiment asked two subjects wearing fNIRS caps to watch the same image together under two conditions as the binary label $H \in \{0, 1\}$, where $H = 1$ corresponds to a handholding condition, and $H = 0$ indicates not-handholding. The fNIRS data, denoted as $\mathcal{D}_1$ and $\mathcal{D}_2$ for each participant, are obtained while simultaneously engaging with the visual stimulus. The classifier aims to predict the label and capture the subtle physiological synchronization and disparities influenced by the interpersonal touch encoded in the hemodynamic responses as measured by fNIRS.
This task extends upon the emerging discourse in social neuroscience regarding the neural underpinnings of human physical interactions as explored in Section \ref{sec:Related_Work_Human_Interactions_Brain_Decoding}. Our approach leverages a paired-sample framework to enhance the classification of interaction states by incorporating both within- and between-subject variations. This method advances our understanding of the neural connectivity patterns associated with social touch. Specifically, within-subject variation involves extracting patterns using the model’s encoder, which processes each subject's data separately. These patterns are then combined with those from paired subjects to analyze between-subject variations, ultimately making the final decision. 
\subsection{Fuzzy Attention Layer}

\subsubsection{Proposition 1: Self-Attention as the Cosine Component of a Fuzzy Set}

While cosine similarity-based scoring is effective in NLP tasks, it may prove inefficient for sparse data patterns. Cosine similarity assigns high scores when the correct orientation is identified and low scores for other orientations. In contrast, the fuzzy distance matrix displays high values in specific neighborhoods and low values in remote areas, offering a distinct advantage in capturing complex patterns.

From the TSK model, the firing strength of a fuzzy rule, denoted as $\overline{f_r}(\mathbf{x})$, can be expressed as follows:

\begin{align}
    \overline{f_r}(\mathbf{x})  
    &= \frac{\mu_r(\mathbf{x})}{\sum_{i=1}^R \mu_i} \nonumber\\
    &= \frac{\exp\left(-\sum_{d=1}^D \frac{(x_{i,d} - m_{r,d})^2}{2\sigma_{r,d}^2}\right)}{\sum_{i=1}^R \exp\left(-\sum_{d=1}^D \frac{(x_{i,d} - m_{r,d})^2}{2\sigma_{r,d}^2}\right)} \nonumber\\
    &= \mathrm{softmax}_{i,r}\left(-\sum_{d=1}^D \frac{(x_{i,d} - m_{r,d})^2}{2\sigma_{r,d}^2}\right) \nonumber\\
    &= \mathrm{softmax}_{i,r}\left(-\sum_{d=1}^D \frac{x_{i,d}^2 + m_{r,d}^2 - 2x_{i,d}m_{r,d}}{2\sigma_{r,d}^2}\right) \nonumber\\
    &= \mathrm{softmax}_{i,r}\left(-\sum_{d=1}^D \frac{x_{i,d}^2 + m_{r,d}^2}{2\sigma_{r,d}^2} + \sum_{d=1}^D \frac{x_{i,d}m_{r,d}}{\sigma_{r,d}^2}\right) \label{eq.fuzzy-atten}
\end{align}

where 
$x_d$ is the input at dimension $d$, $m_{r,d}$ and $m_{r,d}$ represent the centers and widths of the fuzzy rule at dimension $d$.The parameters $m_{r,d}$ and $\sigma_{r,d}$ are learnable and will evolve as training proceeds. Similarly to self-attention, the formula for the fuzzy attention weights can be written as:

\begin{align}
    \text{Attention}(\mathbf{i,j})&= \text{softmax}_{i,j}\left(\frac{\mathbf{Q_iK_j}^T}{\sqrt{d_k}}\right) \nonumber\\
    &=\text{softmax}_{i,j}\left(\sum_{d=1}^D\frac{{x}_{i,d}{k}_{j,d}}{\sqrt{d_k}} \right) \label{eq.self-atten}.
\end{align}


Here, each $x_{i}$ and $k_{j}$ are vectors in the embedded dimension, and their dot product is a scalar.
By comparing equation \ref{eq.self-atten} with \ref{eq.fuzzy-atten}, we observe that the fuzzy firing strength incorporates a term similar to the dot-product self-attention  similarity. The differences between the TSK-type firing strength and dot-product self-attention are:
\begin{enumerate}
    \item TSK not only considers the directional similarity from the dot product but also incorporates the mode of the query and key (fuzzy set).
    \item In self-attention, the length of the key always equals the length of the query, representing the input sequence positions. In contrast, with TSK, the length of keys, represented by fuzzy center $m$, depends on the number of predefined rules.

    \item Self-attention from $i$ to $j$ indicates token $j$'s focus on $i$, while Fuzzy Attention Layer from $i$ to $r$ indicates rule $r$'s focus on token $i$'s.
\end{enumerate}

\begin{figure}[h]
  \centering
  \includegraphics[width=0.95\columnwidth]{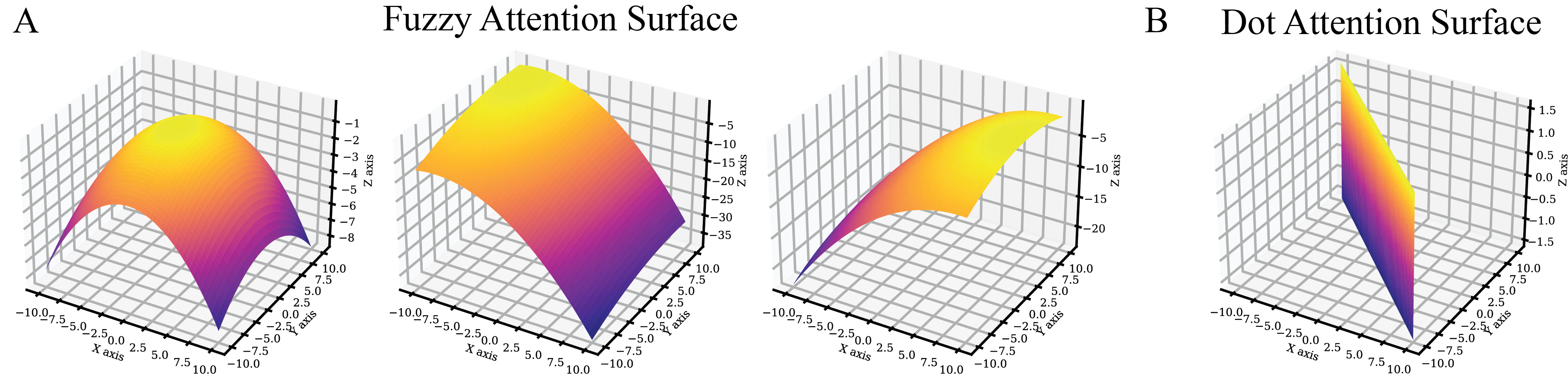}
  \caption{\textbf{A}: The 3 random-selected examples of fuzzy attention score. \textbf{B}: The random-selected dot attention score}
  \label{fig:attention_example}
\end{figure}

With such design, the flexibility of the length of the keys in Fuzzy Attention Layer enables the modeling of input patterns, allowing for analysis and learning of key information, and includes the mode as a contribution to attention. Fig. \ref{fig:attention_example} provides an intuitive illustration of the differences between fuzzy attention(A) and dot-product attention scores(B) based on 2-dimensional inputs. The graphs reveal distinct characteristics of these attention mechanisms. In fuzzy attention, the surfaces exhibit convex shapes, indicating the presence of a maximum scored point. This point offers valuable insights into the patterns learned by the attention model. Conversely, dot-product attention scores form flat planes. When the input query aligns with the key, the attention scales infinitely with the query value. While this orientation-based scoring may be effective in NLP tasks, it proves inefficient for sparse data patterns. Putting simply, cosine similarity yields high scores when the correct orientation is identified and low scores for other orientations. In contrast, the fuzzy distance matrix exhibits high values in certain neighborhoods and low values in remote areas.

\subsubsection{Proposition 2: Reversibility of Queries in Single-Layer Linear Networks}
This section demonstrates that the input to a single-layer linear layer (projector) can be uniquely determined from its output (query), provided that the weight matrix \( W \) is invertible.

The function of the projector can be described by the equation:
\begin{equation}
y = Wx + b
\end{equation}
where \( y \) is the output vector, \( x \) is the input vector, \( W \) is the weight matrix, and \( b \) is the bias vector.

Given the output \( y \), the aim is to find the corresponding input \( x \). We start by rearranging the equation to isolate the term involving \( x \):
\begin{equation}
Wx = y - b
\end{equation}

If the weight matrix \( W \) is invertible, we can apply its inverse to both sides of the equation:
\begin{equation}
x = W^{-1}(y - b)
\end{equation}

This demonstrates that the input \( x \) can be computed directly from the output \( y \) as long as \( W \) is invertible.

However, in cases where \( W \) is not a square matrix or is singular (i.e., not invertible), the Moore-Penrose pseudoinverse \( W^{+} \) is used to compute a least-squares solution:
\begin{equation}
x = W^{+}(y - b)
\end{equation}

The Moore-Penrose pseudoinverse is defined using the Singular Value Decomposition (SVD) of \( W \):
\begin{equation}
W = U \Sigma V^T
\end{equation}
where \( U \) and \( V \) are orthogonal matrices, and \( \Sigma \) is a diagonal matrix of singular values.

The pseudoinverse \( W^{+} \) is computed as:
\begin{equation}
W^{+} = V \Sigma^{+} U^T
\end{equation}
where \( \Sigma^{+} \) is obtained by inverting each non-zero singular value in \( \Sigma \) and transposing the matrix.

This method ensures that if the input data and output query have same dimension, the reconstructed input is same as the original input. 

In summary, the reversibility of queries in a single-layer linear model can be demonstrated under the conditions where the weight matrix \( W \) is either invertible or appropriately handled via pseudoinversion.

\subsection{Fuzzy Transformer for fNIRS modeling}
Our model comprises two main components: the feature extractor and the classification head. The feature extractor follows the structure of a Transformer Encoder, where the attention layers are replaced by a Fuzzy Attention Layer. As depicted in \revised{Fig.} \ref{fig:Sample_Explain}B, this extractor is utilized for both participants in a pair. The output embedding from the encoder is averaged in the second dimension $S$ and concatenated in the final dimension $C$. The shape of the concatenated vector is $2C$. Then, this concatenated vector is processed by the classification head, and the purpose of this concatenation is to combine information across multiple channels and from two subjects and make the final decision. The classification head is a 2-layer Multi-Layer Perception (MLP) model with a ReLU (Rectified Linear Unit) activation function. The output of the classification head consists of two nodes. This integration helps the model to capture and leverage the diverse signal patterns, enhancing its ability to differentiate between two conditions, thereby improving the overall performance and accuracy. The operations of the other basic layers in our encoder follow the standard Transformer architecture, with details available in \cite{NIPS2017_3f5ee243}.

Following equations \ref{eq.fuzzy-atten}, our proposed Fuzzy Attention Layer the input sequence with:

\begin{equation}
    \text{FuzzyAttention}(x) = \text{softmax}\left(\frac{(Q(x) - m)^2}{2\sigma^2}\right)u
\end{equation}
where the $Q(x)$ s the linear projection from the input sequence, and $m$ and $\sigma$ are the Gaussian membership parameters, and $u$ the consequent of the fuzzy rule. From the attention view, the fuzzy attention score replaces the attention score matrix from the dot product similarity to an L2 distance weighted by $\sigma$. This attention score has been observed to be beneficial in fNIRS data modeling.In this paper, the fuzzy rule count is set to 10, based on the results of the ablation study discussed in Section \ref{subsec:eval_rule_vs_prauc}.

This paper defines the objective function as a classification task to determine whether individuals are holding hands. We use cross-entropy loss, which is defined as:

\begin{equation}
\text{loss}(y_{o,c}, p_{o,c}) = -\sum_{c=1}^{M} y_{o,c} \log(p_{o,c}),
\end{equation}

where $y_{o,c}$ represents the true label, $p_{o,c}$ is the predicted probability for class $c$, and $M$ is the total number of classes in the classification problem.

\section{Experiments and Results}

\begin{table*}[ht]
\centering
\caption{Performance comparison of Fuzzy Attention Layer-based transformer against LSTM and vanilla transformer. The evaluation is conducted on two datasets with several matrices listed. * indicate significant differences between our proposed model and others, as assessed by t-tests (\textsuperscript{*}$p < 0.05$,\textsuperscript{**}$p < 0.01$, \textsuperscript{***}$p < 0.001$ \textsuperscript{.}$p<0.1$)}
\label{tab:comparation}
\setlength{\tabcolsep}{1pt}
\renewcommand{\arraystretch}{1.5}
\begin{tabular}{lllllllll}
\toprule
\textbf{Dataset} & \textbf{Structure} & \textbf{Model} & \textbf{Accuracy} & \textbf{Recall} & \textbf{Precision} & \textbf{F1} & \textbf{ROC\_AUC} & \textbf{PR\_AUC} \\ \hline
\multirow{6}{*}{Recognition} & \multirow{3}{*}{channel-first}  & Fuzzy & 67.71\% (0.96\%) & 70.50\% (4.15\%) & 68.98\% (1.78\%) & 69.64\% (1.49\%) & 67.56\% (0.99\%) & 64.11\% (0.90\%) \\
         &           & Transformer & 60.55\% (1.29\%) \textsuperscript{***} & 51.20\% (5.83\%) \textsuperscript{***} & 67.71\% (2.36\%) & 58.08\% (3.47\%) \textsuperscript{***} & 61.30\% (1.17\%) \textsuperscript{***} & 60.80\% (0.87\%) \textsuperscript{***} \\
         &           & LSTM & 55.30\% (0.98\%) \textsuperscript{***} & 59.35\% (7.72\%)\textsuperscript{*} & 58.97\% (1.59\%) \textsuperscript{***} & 58.90\% (3.04\%) \textsuperscript{***} & 54.92\% (1.26\%) \textsuperscript{***} & 57.03\% (0.75\%) \textsuperscript{***} \\
         \cline{2-9}
         & \multirow{3}{*}{time-first} & Fuzzy & 79.85\% (2.75\%) & 82.74\% (2.37\%) & 79.83\% (3.43\%) & 81.23\% (2.34\%) & 79.68\% (2.84\%) & 75.15\% (3.08\%) \\
         &           & Transformer & 76.22\% (1.20\%)\textsuperscript{*} & 77.59\% (3.21\%)\textsuperscript{*} & 78.03\% (0.60\%) & 77.78\% (1.58\%)\textsuperscript{*} & 76.11\% (1.06\%)\textsuperscript{*} & 72.57\% (0.80\%) \\
         &           & LSTM & 65.47\% (2.79\%) \textsuperscript{***} & 70.06\% (3.29\%) \textsuperscript{***} & 64.02\% (1.89\%) \textsuperscript{***} & 66.81\% (0.89\%) \textsuperscript{***} & 62.32\% (2.59\%) \textsuperscript{***} & 58.67\% (4.14\%) \textsuperscript{***} \\
\midrule
\multirow{6}{*}{Rating} & \multirow{3}{*}{channel-first} & Fuzzy & 66.98\% (0.21\%) & 68.14\% (3.48\%) & 69.76\% (1.22\%) & 68.88\% (1.16\%) & 66.89\% (0.23\%) & 64.62\% (0.32\%) \\
         &           & Transformer & 57.36\% (1.75\%) \textsuperscript{***} & 57.49\% (4.77\%)\textsuperscript{**} & 60.04\% (2.42\%) \textsuperscript{***} & 58.60\% (2.09\%) \textsuperscript{***} & 57.35\% (1.88\%) \textsuperscript{***} & 56.85\% (1.22\%) \textsuperscript{***} \\
         &           & LSTM & 64.03\% (0.93\%) \textsuperscript{***} & 69.65\% (3.91\%) & 65.62\% (1.56\%)\textsuperscript{**} & 67.50\% (1.34\%) & 63.58\% (1.07\%) \textsuperscript{***} & 61.97\% (0.82\%) \textsuperscript{***} \\
         \cline{2-9}
         & \multirow{3}{*}{time-first} & Fuzzy & 78.01\% (0.70\%) & 80.49\% (2.41\%) & 79.00\% (0.92\%) & 79.71\% (0.94\%) & 77.81\% (0.63\%) & 74.05\% (0.56\%) \\
         &           & Transformer & 74.21\% (3.96\%)\textsuperscript{.} & 78.78\% (4.30\%) & 74.21\% (4.97\%)\textsuperscript{.} & 76.31\% (3.15\%)\textsuperscript{*} & 73.96\% (4.11\%)\textsuperscript{.} & 69.62\% (3.89\%)\textsuperscript{*} \\
         &           & LSTM & 74.56\% (4.03\%)\textsuperscript{.} & 76.36\% (2.21\%)\textsuperscript{*} & 75.02\% (2.59\%)\textsuperscript{*} & 76.57\% (2.42\%)\textsuperscript{*} & 72.82\% (4.46\%)\textsuperscript{*} & 70.04\% (2.24\%)\textsuperscript{**} \\
\bottomrule
\end{tabular}

\end{table*}

\subsection{Dataset}

To evaluate our proposed Fuzzy Attention Layer aimed at decoding brain signals, we implemented two datasets in our study, each with different experimental setups named Picture Recognition and Picture Rating. Both experiments involved pairs of participants who performed tasks under two conditions: with and without handholding. The experiments were structured to minimize interaction by seating participants back-to-back. \revised{Fig.} \ref{fig:Group_rule.png}D illustrates the fNIRS channel locations. Both two datasets have been approved by the Ethical Review of Psychological Research in Henan Province Key Laboratory of Psychology and Behavior(20200702002).

\begin{figure*}
    \centering
    \includegraphics[width=0.8\linewidth]{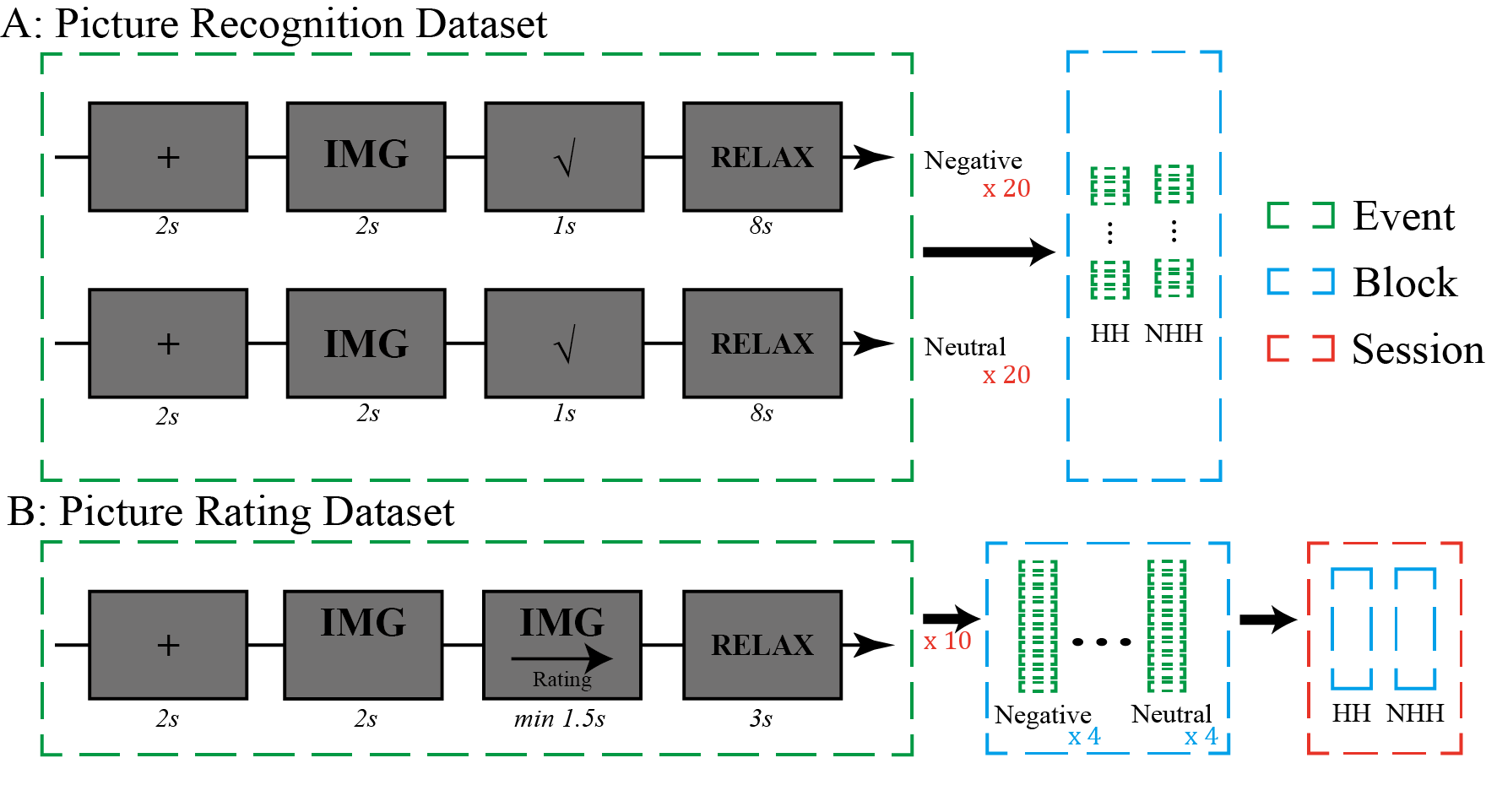}
    \caption{The stimulus diagram of the experiment for two datasets. \textbf{A}: Picture Recognition Dataset and \textbf{B}: Picture Rating Dataset. This figure illustrates how data were collected under the handholding (HH) and non-handholding (NHH) conditions.}
    \label{fig:flowchart}
\end{figure*}

\subsubsection{Picture Recognition}
The study adopted a three-factor mixed experimental design focusing primarily on the variables of relationships(friends vs. strangers), hand-holding (hand-holding vs. non-hand-holding), and image type (negative vs. neutral). In this experiment, participants arrived at the laboratory and were briefed about the study's procedures without revealing the hypotheses. Before the start of the experiment, each participant was isolated in separate rooms to fill out initial questionnaires. Preparation included donning necessary devices. The protocol consisted of three sessions of approximately 5 minutes each for relaxation and briefing, interspersed with two task sessions lasting around 20 minutes each. During the relaxation phases, participants were instructed to remain still, close their eyes, and relax. Task sessions involved emotional image recognition tasks performed both in hand-holding and non-hand-holding conditions, with participants responding to emotional stimuli presented on the screens using a two-button keypad—pressing the left button for negative images and the right for neutral ones. Each image was displayed following a 2-second fixation cross, and the response period was followed by feedback lasting 1 second. Each trial ensured a minimum duration of 11 seconds, extending up to 13 seconds to facilitate relaxation. There are 40 events total for each affective image type in each handholding condition. \revised{The diagram of the experimental stimulus is presented in Fig. \ref{fig:flowchart}A. One example of the collected fNIRS data is shown in Fig. \ref{fig:Sample_Explain}A.}

\subsubsection{Picture Rating}
Similarly, The study also adopted a three-factor mixed experimental design focusing primarily on the variables of relationships(friends vs. strangers), hand-holding (hand-holding vs. non-hand-holding), and image type (negative vs. neutral). The design was organized into a Block design. The protocol was divided into two main blocks based on hand-holding conditions and 16 sub-blocks based on the type of image presented, with each sub-block containing 10 trials of either negative or neutral images. To control for sequence effects, the order of the image type blocks was systematically varied using an ABBA-style arrangement. Each trial commenced with a 2-second presentation of a fixation point, followed by a 2-second solitary display of the image. Participants had 2 seconds to rate the emotional valence (from 1, negative, to 9, positive) and arousal (from 1, calm, to 9, excited). Upon completing ratings, the next trial began. Breaks of 30 minutes and 5 minutes were observed after each image type and hand-holding block, respectively. Participants started with a 5-minute baseline data collection session to measure resting neural activity. Following non-hand-holding tasks, the experiment introduced the hand-holding condition, where participants held hands and repeated the image evaluation task. The study concluded with another 5-minute resting state data collection session to assess any neural changes post-experiment. There are 40 events total for each affective image type in each handholding condition. \revised{The diagram of the experimental stimulus is presented in Fig. \ref{fig:flowchart}B.}

\subsubsection{Participants}
Picture recognition includes 23 dyads of friends (Mean age = 20.156, SD = 2.022) and 26 dyads of strangers (Mean age =19.470, SD = 2.212), and the picture rating task includes 24 dyads of friends (Mean age = 19.786, SD = 1.704) and 23 dyads of strangers (Mean age = 19.932, SD = 1.404).  All subjects are female and recruited through an online registration system. Based on their answer, our potential participants in the friend group had to have known each other for more than six months and consider each other one of their five best friends. Participants in the stranger group were paired randomly to ensure that they had never met before. All participants in this experiment were right-handed, heterosexual, had no history of mental illness, and had normal or corrected vision. Participants were asked to avoid staying up late and drinking alcohol the day before the experiment. They were reassured about the safety of the experiment and provided written informed consent. Participants will receive a compensation of 50 yuan approved by the Ethics Committee of the Henan Provincial Key Laboratory of Psychology and Behavior.

\subsubsection{fNIRS Setup and Preprocessing}
In two experiments, fNIRS signals were simultaneously recorded from two participants at a sampling rate of 7.8125 Hz. The device utilized two wavelengths of near-infrared light, 785 nm and 830 nm, with the frequency adjusted according to the wavelength and channel to avoid crosstalk. Each probe was spaced 30 mm apart. For each participant, 8 sources and 7 detectors were used, resulting in 20 source-detector pair channels (CH) per individual, covering the Prefrontal Cortex (PFC). To ensure signal quality, the Scalp Coupling Index (SCI) was calculated for each channel, and those with an SCI below 0.5 were excluded from further analysis. Next, the raw optical signals were converted into relative oxyhaemoglobin (HbO) and deoxyhaemoglobin (HbR) concentration using the Beer-Lambert law, with a partial pathlength factors (ppf) of 0.1. All fNIRS data were then band-pass filtered between 0.01 Hz and 0.2 Hz. Motion artifacts were removed using a rejection criterion as 120 $\mu$mol/L after epoching, and all data were baseline-corrected using a 2-second baseline for future analysis. All preprocessing was performed using the Python MNE package.



\subsection{Implementation Details}

In our study, the implementation of the neural network model was tailored with specific optimization strategies to enhance training effectiveness and ensure robust convergence. To thoroughly assess the impact of input structure on our model's performance, we have devised two distinct structures aligned with the characteristics of two separate datasets. The data formats are distinguished primarily by their ordering: a time-first format \( (n_{\text{timepoint}}, n_{\text{channel}}) \) and a channel-first format \( (n_{\text{channel}}, n_{\text{timepoint}}) \). Specifically, in the time-first structure, the model employs a center structure analogous to the signal \( (n_{\text{timepoint}}, n_{\text{rule}}) \), thereby emphasizing the temporal dynamics of the data. Conversely, the channel-first structure integrates centers arranged as \( (n_{\text{channel}}, n_{\text{rule}}) \), focusing on the spatial distribution across channels. These structures are intended to explore how different data orientations can influence the learning efficacy and predictive performance of the model. 

Our optimization setup is characterized by using epoch-based iterations, with a total of 800 maximum iterations designated for the training process. We employed the AdamW optimizer, chosen for its effectiveness in managing sparse gradients and incorporating a weight decay of 0.05 for regularization purposes. This optimizer was configured with a dynamically adjusted learning rate, beta coefficients set at 0.9 and 0.95 for the moving averages' exponential decay rates, and a small epsilon value of $1 \times 10^{-8}$ to maintain numerical stability. The learning rate scheduler was another pivotal component of our setup, utilizing the cosine decay learning rate scheduler, which integrates an initial warmup period. This period is crucial as it allows for the gradual escalation of the learning rate, followed by a cosine decay phase, tailoring the learning rate adjustment over the course of training. The effective learning rate, \( lr \), was computed using $lr = \textit{base\_lr} \times \frac{\textit{batch\_size} \times 2}{256}$, and this approach ensures that the learning rate scales appropriately with batch size adjustments, optimizing the training dynamics to accommodate our computational resources and the specifics of the dataset, thereby aligning with the overall experimental framework of our study.

\subsection{Brain Visualization}
We utilize MNE and MNE-NIRS\cite{10.1117/1.NPh.8.2.025008} in Python to project critical data onto the brain's surface, enhancing our visualization capabilities. Additionally, to facilitate the identification of frontal lobes in the figures, we delineate the borders of these lobes, as illustrated in all 3D brain figures.

\subsection{Evaluation Metrics}
\label{subsec:evaluation_metrics}

To assess the performance of our classifier in determining handholding conditions from fNIRS data, we employed six widely recognized metrics: accuracy, recall, precision, F1 score, area under the receiver operating characteristic curve (ROC AUC), and area under the precision-recall curve (PR AUC). Specifically, PR AUC is a robust indicator of the model's ability to classify the positive class in imbalanced datasets, often in physiological data. These metrics provide a comprehensive evaluation by considering the proportion of correct predictions (accuracy), the model's ability to identify true positives among the positives (recall), and its predictive precision. The F1 score balances precision and recall, while ROC AUC and PR AUC provide performance summaries at various threshold settings.

As depicted in Table \ref{tab:comparation}, we show model performance across two dataset arrangements and the statistical results, each with the channel-first and time-first structures, to investigate the effect of input structuring on the classification outcomes. The dimensional reversal aims to explore the robustness of the model against variations in the temporal-spatial data arrangement and explore the explanation capacity of the Fuzzy Attention Layer.

Given that this study pioneers the classification of handholding conditions using fNIRS data, we establish a baseline with bidirectional LSTM and Transformer models, which are state-of-the-art architectures known for their efficacy in sequence modeling tasks. To validate the efficacy of our proposed approach, we performed an independent t-test to compare it with other models. Each model was tested by training it five times using different random seeds. The Fuzzy Attention Layer-based model achieves the Top 1 in PR AUC metrics in Image Recognition, channel-first, Image Rating channel- and time-first experiments ($p < 0.05$).

\begin{figure}
    \centering
    \includegraphics[width=0.9\columnwidth]{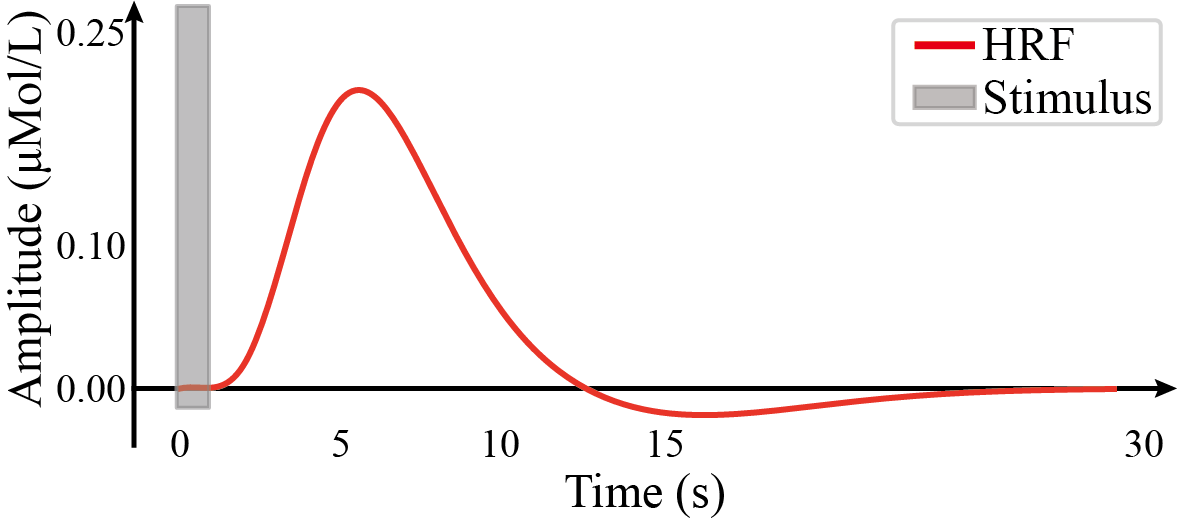}
    \caption{\textbf{HRF}, modelled by the standard SPM method. The figure illustrates the response over time to a stimulus of 1-second duration starting at 0 seconds. The response is depicted as a continuous curve, showcasing the typical biphasic pattern of the HRF, with an initial rise peaking shortly after stimulus onset followed by an undershoot. The shaded area represents the stimulus's duration, highlighting the neural activation's temporal context.}

    \label{fig:HRF}
\end{figure}

\subsection{Ablation Study}

\subsubsection{Exploring the Impact of Fuzzy Attention Layer Replacements within Transformer Encoders}

In our pursuit to understand the impact of the Fuzzy Attention Layers in the Transformer Encoder, we use the Picture Recognition dataset and replace the attention layers in two different structures (time-first and channel-first) in a 3-layer model as a demonstration. This experimental design allows us to systematically explore the effects of layer-specific modifications within our Transformer encoder architecture. We evaluate the performance of each structure containing replaced layers in various combinations (individual layers, two layers, and all three layers together, total 6 combinations). The performance metrics are calculated for each experiment, providing a comprehensive view of the impact of these replacements. The results, as detailed in Table \ref{tab:replacedlayers}, show that different structures respond uniquely to replacing attention layers. For example, replacing all three layers in the channel-first structure significantly enhances both the model's accuracy and recall, suggesting a cumulative benefit of layer replacement in complex encoding scenarios. Conversely, in the time-first structure, the selective replacement of individual layers appears more beneficial, particularly for precision and F1 scores. These findings underscore the nuanced role of Fuzzy Attention Layers in enhancing the Transformer Encoder's performance. By isolating the effects of specific layers, we can better understand the dynamics of attention mechanisms and their contributions to the model's overall effectiveness in handling different structures.

\begin{table*}[h]
\centering
\caption{Table Encoder Metrics in Picture Recognition and Where the attentions are replaced in 3 Layers by Fuzzy Attention Layers}
\label{tab:replacedlayers}
\setlength{\tabcolsep}{3pt}
\renewcommand{\arraystretch}{1.3}
\begin{tabular}{llcccccc}
\toprule
\textbf{Structure} & \textbf{Replaced Layers} & \textbf{accuracy} & \textbf{recall} & \textbf{precision} & \textbf{F1} & \textbf{roc\_auc} & \textbf{pr\_auc} \\
\midrule
\multirow{7}{*}{channel-first} & \#1               & 59.64\% & 58.80\% & 62.36\% & 60.53\% & 59.68\% & 58.35\% \\
                               & \#2               & 56.40\% & 55.01\% & 59.24\% & 57.05\% & 56.48\% & 56.27\% \\
                               & \#3               & 61.79\% & 57.59\% & 65.61\% & 61.34\% & 62.02\% & 60.10\% \\
                               & \#1 \& \#2        & 62.16\% & 57.71\% & 66.09\% & 61.62\% & 62.40\% & 60.40\% \\
                               & \#1 \& \#3        & 60.66\% & 56.27\% & 64.47\% & 60.09\% & 60.91\% & 59.29\% \\
                               & \#2 \& \#3        & 55.51\% & 55.57\% & 58.08\% & 56.80\% & 55.51\% & 55.66\% \\
                               & \#1, \#2 \& \#3   & \textbf{67.42\%} & \textbf{69.39\%} & \textbf{68.92\%} & \textbf{69.15\%} & \textbf{67.31\%} & \textbf{63.94\%} \\
                                                    
\cline{2-8}
\multirow{7}{*}{time-first}    & \#1               & 76.43\% & 78.58\% & \textbf{77.09\%} & 77.83\% & 76.31\% & 71.85\% \\
                               & \#2               & 72.16\% & 84.93\% & 69.19\% & 76.25\% & 71.45\% & 66.69\% \\
                               & \#3               & 73.73\% & 76.67\% & 74.25\% & 75.44\% & 73.56\% & 69.21\% \\
                               & \#1 \& \#2        & 76.59\% & 83.11\% & 75.08\% & 78.89\% & 76.23\% & 71.29\% \\
                               & \#1 \& \#3        & \textbf{78.73\%} & \textbf{85.28\%} & 76.86\% & \textbf{80.85\%} & \textbf{78.37\%} & \textbf{73.29\%} \\
                               & \#2 \& \#3        & 72.50\% & 79.00\% & 71.66\% & 75.15\% & 72.14\% & 67.66\% \\
                               & \#1, \#2 \& \#3   & 77.77\% & 82.30\% & 77.03\% & 79.58\% & 77.52\% & 72.71\% \\
\bottomrule
\end{tabular}
\end{table*}

\subsubsection{Investigating the Rule Count for Model Optimization}
\label{subsec:eval_rule_vs_prauc}

\begin{figure}[ht]
\centering
\includegraphics[width=\linewidth]{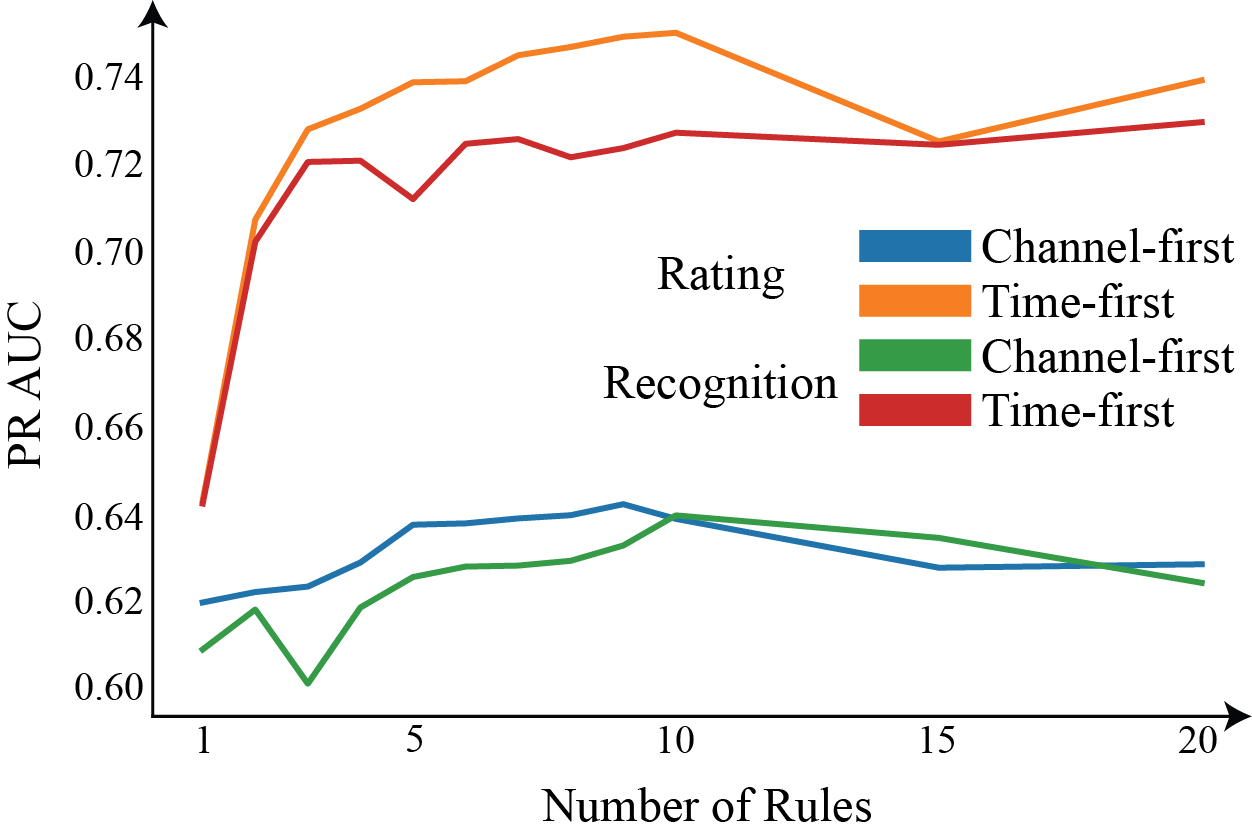}
\caption{Variation of PR AUC with the number of rules indicating initial improvement followed by a performance decline due to model overfitting.}
\label{fig:number_of_rules}
\end{figure}

The efficacy of the model is analyzed with respect to the number of rules employed within the computational framework. \revised{Fig.} \ref{fig:number_of_rules} delineates the relationship between the PR AUC and the rule count, indicating a nuanced dependency on the number of rules. Initially, the PR AUC exhibits an enhancement as the number of rules increases, peaking at approximately ten rules. After that, the model's performance demonstrates a declining trajectory post the apex, characterized by fluctuations in the PR AUC. This downturn signals a saturation point, beyond which further additions of rules contribute to overfitting. The increase of an excessive number of rules escalates the model’s complexity and parameter count, which, although aimed at refining precision on training data, paradoxically impairs its generalization capabilities on novel, unseen datasets. Consequently, the model exhibits diminished efficacy, underscoring the critical balance required in the rule count to maintain optimal performance without succumbing to overfitting.

\subsubsection{Impact of Model Depth on Encoder Performance}
\label{subsec:depth_impact_prauc}

To evaluate the influence of increasing the depth of our model, we systematically varied the number of hidden layers and recorded the changes in model performance. The results of this experiment are depicted in Table \ref{tab:depth}, where the metrics values for models with 1, 2, and 3 layers are shown. Channel-first and time-first structures across three different depth levels. For the Picture Recognition dataset under the channel-first configuration, the encoder shows gradual improvement in metrics as depth increases, peaking at a depth of 3 with an accuracy of 67.42\%, recall of 69.39\%, and an F1 score of 69.15\%. Similarly, under the time-first structure, the best performance is observed at a depth of 2 with precision reaching 76.98\% and F1 score peaking at 81.35\%. In contrast, for the Picture Rating dataset, the encoder achieves the highest accuracy and precision under the time-first structure at depth 2 with values of 79.32\% and 79.29\%, respectively. This table illustrates the significant impact of structure and depth on the encoder's performance, guiding the selection of optimal encoder settings for specific types of data analysis.

\begin{table*}[ht]
\centering
\caption{Table of Encoder Metrics by Dataset, Reversal, and Depth}
\label{tab:depth}
\setlength{\tabcolsep}{3pt}
\renewcommand{\arraystretch}{1.3}
\begin{tabular}{llclcccccc}
\toprule
\textbf{Dataset}        & \textbf{structure}         & \textbf{Depth} & \textbf{accuracy} & \textbf{recall} & \textbf{precision} & \textbf{F1} & \textbf{roc\_auc} & \textbf{pr\_auc} \\
\midrule
\multirow{6}{*}{Recognition}  & \multirow{3}{*}{channel-first} & 1 & 66.44\% & 67.85\% & 68.22\% & 68.03\% & 66.36\% & 63.21\% \\
                        &                           & 2 & 67.20\% & 67.60\% & \textbf{69.32\%} & 68.45\% & 67.18\% & 63.91\% \\
                        &                           & 3 & \textbf{67.42\%} & \textbf{69.39\%} & 68.92\% & \textbf{69.15\%} & \textbf{67.31\%} & \textbf{63.94\%} \\
                        \cline{2-9}
                        & \multirow{3}{*}{time-first} & 1 & 78.37\% & 80.65\% & \textbf{78.76\%} & 79.69\% & 78.24\% & \textbf{73.70\%} \\
                        &                           & 2 & \textbf{79.19\%} & \textbf{86.25\%} & 76.98\% & \textbf{81.35\%} &  \textbf{78.79\%} & 73.63\% \\
                        &                           & 3 & 77.77\% & 82.30\% & 77.03\% & 79.58\% &77.52\% & 72.71\% \\
\midrule
\multirow{6}{*}{Rating} & \multirow{3}{*}{channel-first} & 1 &  \textbf{65.38\%} & \textbf{62.03\%} & 70.07\% & \textbf{65.81\%} & 65.65\% & 63.86\% \\
                        &                           & 2 & 65.18\% & 59.00\% & \textbf{71.22\%} & 64.54\% & \textbf{65.68\%} & \textbf{64.04\%} \\
                        &                           & 3 &65.26\% & 61.07\% & 70.34\% & 65.38\%& 65.60\% & 63.86\% \\
                        \cline{2-9}
                        & \multirow{3}{*}{time-first} & 1 & 76.67\% & 77.92\% & 78.49\% & 78.20\% & 76.57\% & 73.02\% \\
                        &                           & 2 & 78.21\% & 80.11\% & \textbf{79.48\%} & 79.80\% & 78.06\% & 74.36\% \\
                        &                           & 3 & \textbf{79.32\%} & \textbf{83.25\%} & 79.29\% & \textbf{81.22\%} & \textbf{79.01\%} & \textbf{75.00\%} \\
\bottomrule
\end{tabular}
\end{table*}

\section{Discussion}

This section presents a detailed examination of how we identify and interpret neural activity patterns.

\subsection{Fuzzy Set and Feature Distribution}

\begin{figure*}
    \centering
    \includegraphics[width=0.8\linewidth]{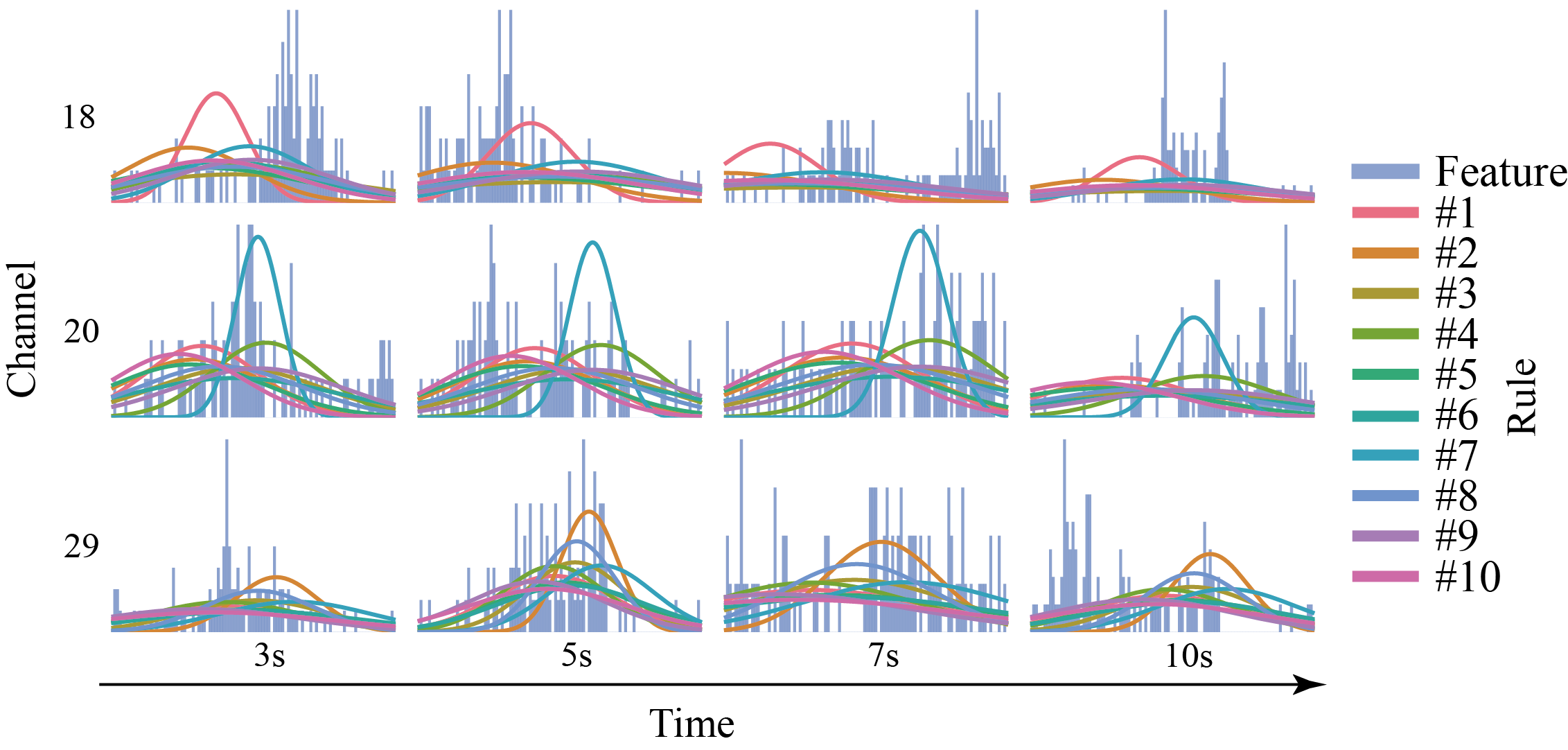}
    \caption{Temporal and Channel-Specific Fuzzy Set and Feature Distributions: This figure illustrates the dynamic distribution of fuzzy membership values across different EEG channels at selected time points (3s, 5s, 7s, 10s). The histograms represent fNIRS data, while the superimposed colored curves correspond to different fuzzy logic rules applied to interpret the neural activity.}
    \label{fig:attention_distribution}
\end{figure*}

The fuzzy set distributions of each fuzzy rule, as depicted in \revised{Fig.} \ref{fig:attention_distribution}, illustrate the temporal dynamics and channel specificity of fuzzy membership functions over multiple experimental conditions. The channels 18, 20, and 29 graphs highlight the variability in attention across different time intervals (3s, 5s, 7s, 10s). Each colored line represents a distinct rule derived from fuzzy logic that interprets the fNIRS data, which is overlaid as blue histograms. 

Notably, the peaks in the data histograms correspond to moments of significant neural activity, while the overlays of fuzzy rules provide an interpretative framework to understand these peaks in the context of cognitive processes. Rule \#1 (red), Rule \#2 (orange), and Rule \#3 (yellow) show a marked variance in their alignment with the actual data peaks, suggesting different cognitive mechanisms being engaged at these times.

For example, in Channel 18 at the 3s and 7s marks, the alignment of Rule \#1 with the data peak indicates a strong correspondence with the expected neural signature of focused attention. Conversely, the broader spread of Rule \#3 around the 10s mark across all channels suggests a more diffused attention state, possibly indicative of a transition between cognitive states.

\revised{These results show the diversity of the learned fuzzy logic-based rules to dissect the complex temporal patterns.}

\subsection{Sample-wised Interpretability Analysis}

To provide an intuitive illustration of how the Fuzzy Attention Layer identifies the handholding condition from fNIRS data, we present \revised{the learned center prototype and firing strength for the example fNIRS data in Fig \ref{fig:Sample_Explain}C and \ref{fig:Sample_Explain}E. To better understand the inference processing, we also present} a demo sample from the top-performing model, where the subject holds a hand.

\begin{figure*}[t]
  \centering
  \begin{minipage}[t]{0.28\linewidth}
    \centering
    \raisebox{-1.1\height}{\includegraphics[width=0.90\linewidth]{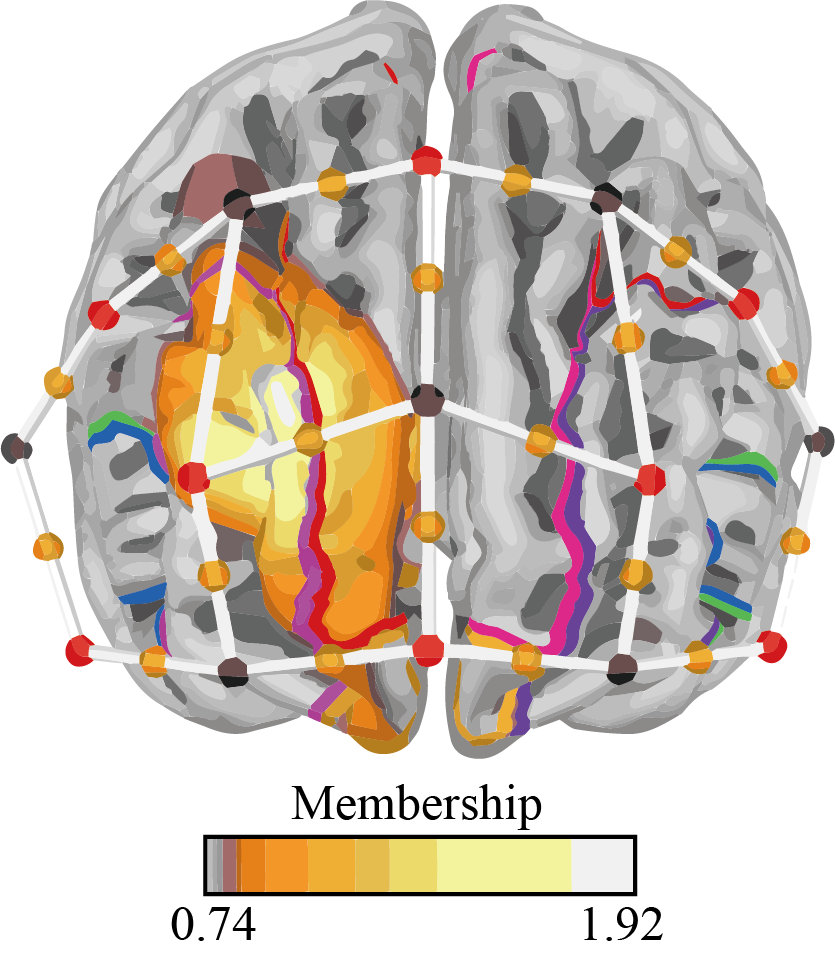}}
    \caption{Visualization of membership values in rule \#10.}
  \end{minipage}\quad
  \begin{minipage}[t]{0.65\linewidth}
    \centering
    \tabcaption{Sample analysis: rule firing strength with top3  channel identified.}
    \renewcommand{\arraystretch}{0.85}
    \begin{tabular}{cccccc}
    \toprule
    Rule & Firing Strength & Top1 & Top2 & Top3 \\
    \midrule
    1  & 0.13 & 5  & 15 & 11 \\
    2  & 0.00 & 8  & 10 & 5  \\
    3  & 0.01 & 7  & 11 & 5  \\
    4  & 0.07 & 39 & 30 & 14 \\
    5  & 0.10 & 11 & 36 & 15 \\
    6  & 0.17 & 5  & 35 & 30 \\
    7  & 0.01 & 33 & 13 & 36 \\
    8  & 0.01 & 13 & 10 & 15 \\
    9  & 0.17 & 13 & 5  & 38 \\
    10 & 0.32 & 11 & 30 & 36 \\
    \bottomrule
    \label{tab:sample_ana}
    \end{tabular}
  \end{minipage}
\end{figure*}

From Table \ref{tab:sample_ana}, we can see that considering border firing strength, the contributions from channels 5 (HbO in the Left ventromedial prefrontal cortex, VmPFC), 11 (HbO in the Left frontopolar area, FPA), and 30 (HbR in the Right dorsolateral prefrontal cortex, dlPFC) are significant. Channel 5 is highlighted by rules \#1, \#6, and \#9; channel 11 is highlighted by rules \#1 and \#10; and channel 30 is highlighted by rules \#6 and \#10. The dominant rule is \#10, where channels 11, 30, and 36 (HbR in the Right VmPFC) contribute the most.

\subsection{Group-level Interpretability Analysis}

To investigate the decision-making mechanism of the rule as a filter, we conducted statistical analyses to assess the differences between two labels under one rule at the group level. We also compared the IBS difference between the labels.

\subsubsection{Channel-wise Analysis}
We selected the model demonstrating the highest performance for the Picture Recognition dataset, which included 10 rules. We chose this top-performing model and used a t-test to analyze the differences in firing strength. The t-values describe the differences between handholding and non-handholding in the PFC, as shown in \revised{Fig.} \ref{fig:Group_rule.png}A. The results indicate how the rules contribute to classification. For example, rules \#1, \#2, \#3, \#5, \#9, and \#10 involve HbO in the right lPFC.

\subsubsection{Time-wise Analysis}

Similarly, in the Picture Recognition dataset, we employed a t-test to evaluate the differences in firing strength between the labels. The t-values illustrating these differences are depicted in \revised{Fig.} \ref{fig:Group_rule.png}B. The results indicate that the crucial time domain is approximately 5 seconds after stimuli presentation in rules \#5, \#6, \#7, and \#8, which is highly related to the \revised{Haemodynamic Response Function (HRF) \cite{SMITH200983}. The typical \revised{simulated} HRF is shown in \revised{Fig.} \ref{fig:HRF}. This adherence to the HRF is critical as HRF-like data patterns exhibit stable temporal characteristics that are optimally suited for analysis by computational models. Particularly, implementing a fuzzy attention layer within our model shows substantial improvements in performance, as it can more effectively learn and adapt to these HRF-like patterns. \revised{Fig.} \ref{fig:Sample_Explain}C also demonstrates the HRF-like pattern that the center has learned to recognize. This capability underscores the model’s enhanced interpretability and robustness when handling time-first fNIRS data.}

Rule \#5, in particular, shows the most significant difference between handholding and non-handholding conditions(\textit{p} < 0.05). \revised{Fig.} \ref{fig:Group_rule.png}C displays the center prototype in PFC for both oxygenated (HbO) and deoxygenated hemoglobin (HbR), and we also observe the left prefrontal cortex (lPFC)'s contribution in the fuzzy centers noted in the Channel-wise model observation.

\subsubsection{IBS Analysis}
To investigate IBS during the evaluation of affective images in the context of interpersonal touch, and to examine the role of IBS in alleviating negative emotional responses, we calculated various (de-)synchrony metrics. These metrics included Pearson correlation, cosine similarity, and Euclidean distance between the brain signal embeddings of two participants, $\mathbf{e}_1$ and $\mathbf{e}_2$, extracted using the Fuzzy Attention Layer. $\mathbf{e}_1$ and $\mathbf{e}_2$ can be described as:

\begin{equation}
\mathbf{e}_{1(2)} = \text{FuzzyAttentionLayer}(\mathcal{D}_{1(2)})
\end{equation}

The results indicate that both correlation and cosine similarity metrics showed significantly higher IBS in the handholding condition compared to the non-handholding condition ($\textit{p} < 0.05$), as shown in \revised{Fig.} \ref{fig:Group_rule.png}E.

\subsection{Result Explanation}

The PFC encompasses several key areas integral to cognitive and emotional regulation. The left PFC plays a pivotal role in cognitive control, crucial for executive functions such as planning and decision-making \cite{NGUYEN2021118599}. vmPFC is crucial in regulating the hypothalamo-pituitary-adrenal (HPA) axis's response to emotional stress, thereby playing a significant role in emotional resilience and stress management \cite{Radley2006RegionalDO}. The function of FPA is less well understood, but it is hypothesized to participate in cognitive branching, a process essential for complex problem-solving and strategic thinking \cite{PFA2001Semendeferi, BrainofLB1}.

Moreover, dlPFC is implicated in regulating emotional valence and influencing our experience and interpretation of emotions. The vmPFC may aid in the extinction of arousal from emotional stimuli, which is vital for emotional regulation and recovery \cite{Nejati2021}. Furthermore, our study emphasizes the critical role of IBS in emotional communication. Research by Hertenstein demonstrates that touch not only conveys but also accurately interprets emotions; our findings support this by showing that handholding significantly enhances IBS, which is linked to pain relief \cite{Hertenstein2006TouchCD, GoldsteinPainreduction}.

Expanding on these regional functions within the PFC, our research delves into the specific effects of handholding on these neural pathways. Handholding seems to modulate activity within the lPFC and dlPFC, likely enhancing cognitive and emotional regulation during stress. This modulation may contribute to a calming effect, evidenced by the observed reduction in stress markers and reported increases in emotional comfort. The observed synchronization of brain activities between partners during handholding likely helps to align emotional states, fostering mutual empathy and support. This phenomenon highlights the significant role of simple physical contact in enhancing psychological resilience and stress-coping mechanisms. Supported by prior research linking physical touch to better health outcomes and emotional stability \cite{FIELD2010367, fieldTouch2001}, these insights suggest that interpersonal touch, particularly handholding, could be effectively incorporated into therapeutic practices to boost emotional regulation and strengthen interpersonal bonds.

\subsection{Error Analysis}

\begin{figure*}
    \centering
    \includegraphics[width=0.8\linewidth]{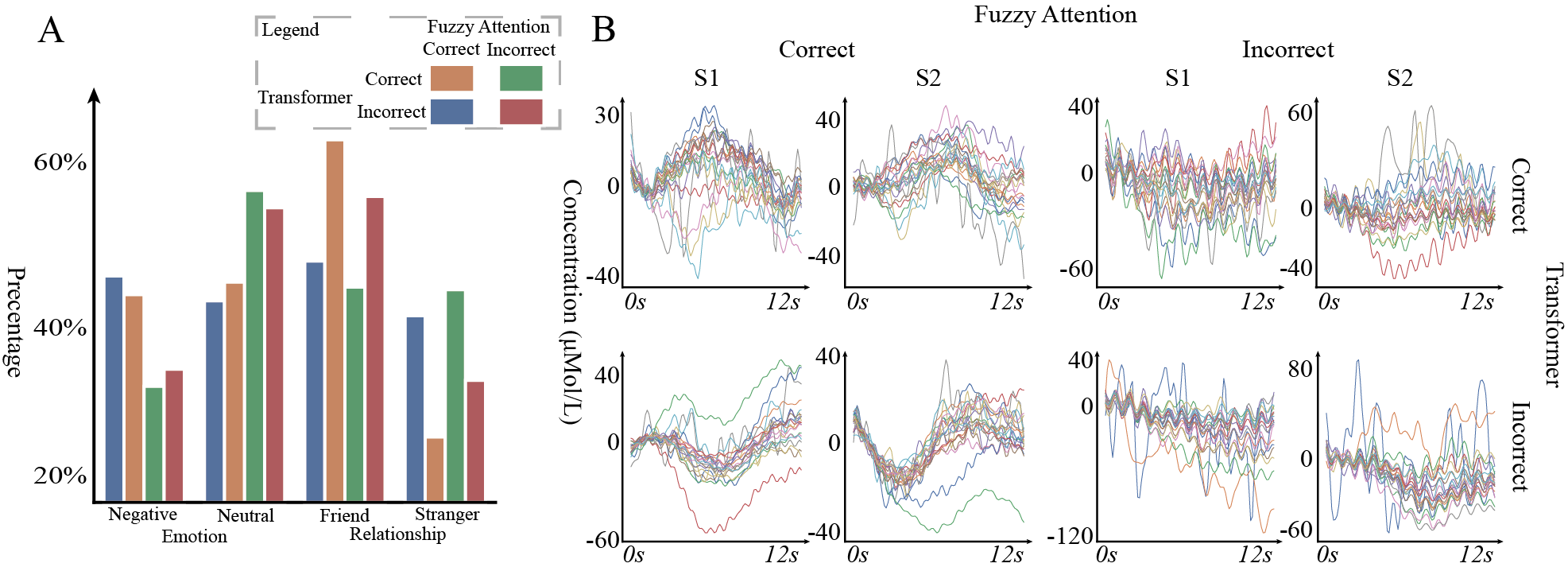}
    \caption{Error Analysis. \textbf{A}: The percentage of model predictions across different data types using the Fuzzy Attention model and Vanilla Attention model. \textbf{B}: Representative samples for the four possible model prediction conditions ($2x2$ grid).}
    \label{fig:ErrorAnalysis}
\end{figure*}

In this section, we analyze the prediction performance of the two best-performing models in our test: the Fuzzy Attention model and the Vanilla Attention model. We evaluate their performance across different data types (emotional image: negative vs. neutral; pair-subject relationship: stranger vs. friend), as shown in \revised{Fig.} \ref{fig:ErrorAnalysis}A. The results indicate that the Fuzzy Attention model demonstrates a lower percentage of incorrect predictions when processing negative emotional images compared to neutral ones, while it shows minimal variance across different relationship categories. In contrast, the Vanilla Attention model exhibits more uniform performance across both data types.

The superior performance of the Fuzzy Attention model, particularly with negative stimuli, aligns with findings in neuroimaging studies that highlight how negative emotional stimuli elicit distinct neural and physiological responses, especially in regions such as the amygdala and prefrontal cortex, which are responsible for emotion regulation and attention control \cite{Hamilton2012FunctionalNO, Marekov2016BrainAA, Wiener2020BloodPR, OkonSinger2014NeuralCO}. This complex interplay between emotion, attention, and physiological responses emphasizes the model's ability to capture the heightened brain blood response that typically accompanies negative emotional stimuli. 

\revised{Fig.} \ref{fig:ErrorAnalysis}B further analyzes typical samples, comparing the patterns identified by the Fuzzy Attention and Vanilla Attention models. It is evident that when the data aligns closely with HRF,
the Fuzzy Attention model successfully predicts the outcome. However, when the data fails to elicit a standard HRF response, possibly due to atypical stimuli, the Fuzzy Attention model struggles to make accurate predictions.

In conclusion, the Fuzzy Attention model excels at recognizing stable neural patterns, particularly those associated with stable responses, such as watching affective stimuli, allowing it to outperform the Vanilla Attention model in scenarios where distinct physiological signals are present.

\subsection{Future Work}
Our proposed model shows a potential fuzzy-rule-based approach to understanding the neural patterns in social neuroscience. 
While this study has focused on exploring and uncovering neuroscience and psychological theories using fuzzy logic-based models, future work could include optimizing the design of fuzzy centers and automating the decision-making process for the number of fuzzy rules. Additionally, improving the speed of training is essential for enhancing the practical applicability of the model. 
By addressing these challenges, researchers can enhance the model's practical applicability and potentially expand its use in real-time social neuroscience studies, paving the way for innovations in therapeutic strategies and human-machine interaction. 
Future studies should also consider conducting longitudinal research to evaluate the enduring effects of social interactions on neural dynamics, which could further validate and refine the contributions of this research.

\section{Conclusion}
In this work, we introduce a new attention layer, the Fuzzy Attention Layer, to identify interpretable patterns of neural activity, which are crucial for detailed analysis and understanding of the decision-making process of the neural network for individual cases. This method enhances the correlation modeling ability for fNIRS data. Our paper demonstrates that the Fuzzy Attention Layer outperforms vanilla self-attention in modeling boundary relations. Additionally, the transformer model with a Fuzzy Attention Layer achieves higher approximation accuracy than the vanilla transformer. This finding is validated with two fNIRS datasets, where relevant neural activity patterns are identified. Given the potential of the Fuzzy Attention Layer, its application extends to various neural data types, such as EEG or fMRI, which can significantly enhance neuroscience and BCI technology by decoding and interpreting neural patterns in a human-understandable form. Our contribution sets the stage for substantial advancements in comprehending the brain's intricate mechanisms and improving neurotechnology. Looking forward, the future scope of this method includes refining the designs of fuzzy centers and the number of rules, accelerating the training process, and implementing the model in real-time social neuroscience settings to explore neuroscience and psychological theories, develop sophisticated therapeutic strategies, and elevate human-machine interactions.


\section{Limitations}
This study has identified several limitations. First, the computational speed is notably low due to the reliance on the Transformer architecture. A potential remedy is to reduce the model size to enhance processing efficiency. Second, the presence of similar rules contributes to redundancies in model parameters. Optimizing for rule diversity could alleviate this issue and improve model performance.

\bibliographystyle{IEEEtran}
\bibliography{ref}
\end{document}